\documentclass{article}

\usepackage{microtype}
\usepackage{graphicx}
\usepackage{subcaption}
\usepackage{booktabs}
\usepackage{tcolorbox}
\usepackage{wrapfig}
\usepackage{hyperref}
\usepackage{multirow}
\usepackage{colortbl}

\definecolor{brickred}{rgb}{0.6, 0, 0} 

\usepackage[preprint]{icml2026}
\renewcommand{\headrulewidth}{0pt} 
\usepackage{amsmath}
\usepackage{amssymb}
\usepackage{mathtools}
\usepackage{amsthm}
\usepackage{enumitem}
\usepackage{wrapfig}

\usepackage{xcolor}
\usepackage{tcolorbox} 
\definecolor{acadred}{RGB}{160, 20, 20} 
\definecolor{acadredbg}{RGB}{250, 240, 240} 

\hypersetup{
    colorlinks=true,
    linkcolor=acadred,  
    citecolor=acadred,  
    urlcolor=acadred    
}

\newtcolorbox{researchquestion}{
    colback=acadredbg,       
    colframe=acadred,        
    boxrule=0.8pt,           
    arc=2pt,                 
    width=\linewidth,        
    left=4pt, right=4pt,     
    top=4pt, bottom=4pt,     
    before skip=24pt,         
    after skip=24pt,          
    fontupper=\sffamily\bfseries\centering 
}

\newtcolorbox{findings}{
    colback=acadredbg,       
    colframe=acadred,        
    boxrule=0.8pt,           
    arc=2pt,                 
    width=\linewidth,        
    left=4pt, right=4pt,     
    top=4pt, bottom=4pt,     
    before skip=8pt,         
    after skip=8pt,          
    fontupper=\centering 
}

\newtcolorbox{insightbox}[1]{
    colback=white,          
    colframe=acadred,       
    colbacktitle=acadredbg, 
    coltitle=acadred,       
    fonttitle=\bfseries,    
    title={#1},             
    boxrule=0.6pt,
    arc=2pt,
    attach boxed title to top left={yshift=-2mm, xshift=2mm}, 
    enhanced,               
    left=4pt, right=4pt, top=8pt, bottom=4pt
}

\newtcolorbox{definitionbox}{
    blanker,                
    borderline west={2pt}{0pt}{acadred}, 
    colback=acadredbg,      
    left=8pt, right=8pt, top=6pt, bottom=6pt,
    arc=0pt
}

\DeclareCaptionFont{acadredfont}{\bfseries}
\usepackage{sectsty} 

\usepackage{sectsty}
\sectionfont{\sffamily\bfseries} 
\subsectionfont{\sffamily\bfseries}

\usepackage[capitalize,noabbrev]{cleveref}
\usepackage{lmodern}
\usepackage{roboto}

\theoremstyle{plain}

\theoremstyle{definition}

\theoremstyle{remark}

\usepackage[textsize=tiny]{todonotes}

\icmltitlerunning{}

\makeatletter
\renewcommand{\paragraph}{\@startsection{paragraph}{4}{\z@}
    {-0.2ex plus -0.5ex minus -1ex} 
    {-1em}                            
    {\sffamily\bfseries}}
\makeatother

\begin{document}

\makeatletter
\icml@noticeprintedtrue 
\makeatother

\begin{tcolorbox}[
    colback=acadredbg,
    colframe=acadredbg,
    arc=4mm,
    boxsep=3mm, 
    width=\linewidth
]
    \vspace{3mm}
    \begin{center}
        {
            \sffamily \bfseries \fontsize{18}{20} \selectfont
            \color{acadred} Evaluating Parameter Efficient Methods for RLVR
        }
        \par
        \vspace{4mm}
        {
            \sffamily 
            \fontsize{10}{3}
            \selectfont
             \textbf{Qingyu Yin}$^{1*}$, 
             \textbf{Yulun Wu}$^{1*}$, 
             \textbf{Zhennan Shen}$^{2*}$,
             \textbf{Sunbowen Li}$^{3}$, 
             \textbf{Zhilin Wang}$^{4}$, 
             \textbf{Yanshu Li}$^{5}$, 
             \\ \textbf{Chak Tou Leong}$^{6}$,
             \textbf{Jiale Kang}, 
             \textbf{Jinjin Gu}$^{7}$
        }
        \vspace{2mm} \\
        {
            \fontsize{10}{14}
            \selectfont
            \color{black!80}
            $^1$Zhejiang University, 
            $^2$HKUST, 
            $^3$WUST, 
            $^4$USTC,
            $^5$Brown University, \\
            $^6$Hong Kong Polytechnic University, 
            $^7$INSAIT \\
            $^*$Equal contribution. 
            Correspondence to: {\color{acadred}qingyu.yin@zju.edu.cn.}
        }
    \end{center}
    \vspace{3mm}
    
    \noindent
We systematically evaluate Parameter-Efficient Fine-Tuning (PEFT) methods under the paradigm of Reinforcement Learning with Verifiable Rewards (RLVR). RLVR incentivizes language models to enhance their reasoning capabilities through verifiable feedback; however, while methods like LoRA are commonly used, the optimal PEFT architecture for RLVR remains unidentified. In this work, we conduct the first comprehensive evaluation of over 12 PEFT methodologies across the DeepSeek-R1-Distill families on mathematical reasoning benchmarks. Our empirical results challenge the default adoption of standard LoRA with three main findings. First, we demonstrate that structural variants, such as DoRA, AdaLoRA, and MiSS, consistently outperform LoRA. Second, we uncover a spectral collapse phenomenon in SVD-informed initialization strategies (\textit{e.g.,} PiSSA, MiLoRA), attributing their failure to a fundamental misalignment between principal-component updates and RL optimization. Furthermore, our ablations reveal that extreme parameter reduction (\textit{e.g.,} VeRA, Rank-1) severely bottlenecks reasoning capacity. We further conduct ablation studies and scaling experiments to validate our findings. This work provides a definitive guide for advocating for more exploration for parameter-efficient RL methods.
\vspace{0.2cm}

\begin{center}
    \textbf{\sffamily Code:} \href{https://github.com/MikaStars39/PeRL}{\texttt{PeRL}} 
  \quad 
  \textbf{\sffamily Checkpoints:} \href{https://huggingface.co/MikaStars39/PeRL}{\texttt{HuggingFace}}
  \quad
  \textbf{\sffamily Logs:} \href{https://wandb.ai/mikastars-zhejiang-university/PeRL_logs}{\texttt{Wandb}}
\end{center}

\end{tcolorbox}

\makeatletter
\global\icml@noticeprintedtrue 
\makeatother

\vspace{0.2cm} 



\begin{figure}[ht]
    \centering
    \vspace{-0.3cm}
    \begin{subfigure}[t]{0.47\linewidth} 
        \centering
        \includegraphics[width=1\textwidth]{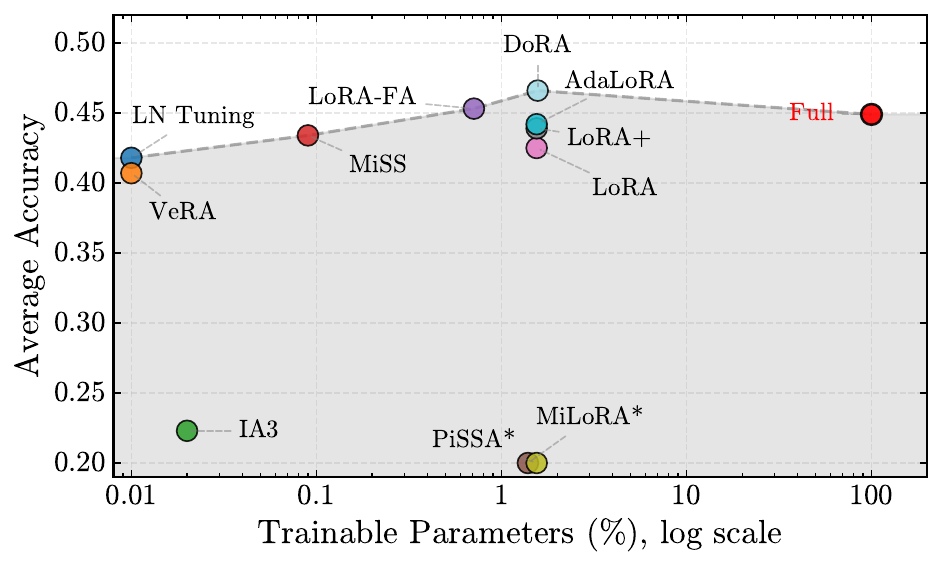} 
        \label{fig:reward_b}
        \vspace{-0.6cm}
    \end{subfigure}
    \hfill
    \begin{subfigure}[t]{0.49\linewidth} 
    \centering
    \includegraphics[width=1\textwidth]{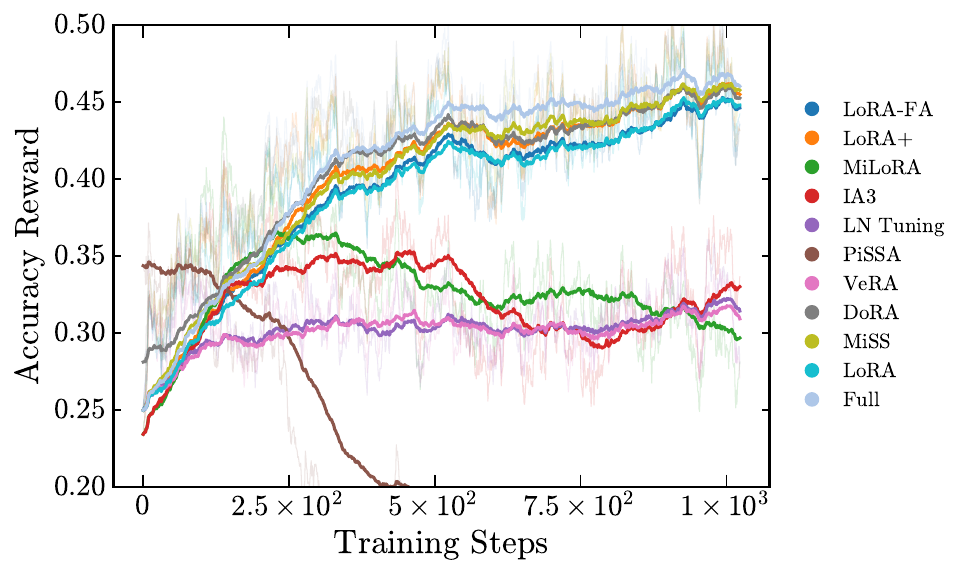} 
    \label{fig:reward_a}
    \vspace{-0.6cm}
    \end{subfigure}
    \label{fig:reward_bench}
    \caption{\textbf{Left:} Comparison of average accuracy \textit{vs.} percentage of trainable parameters (log scale) for various parameter efficient methods under our RLVR evaluations. The shaded area represents the performance frontier. \textbf{Right:} Training dynamics showing accuracy reward over training steps for different methods.}
    \vspace{-0.5cm}
\end{figure}

\section{Introduction}
Large Language Models (LLMs) ~\cite{vaswani2017attention,brown2020language} have demonstrated remarkable proficiency in complex reasoning tasks, particularly within mathematical and scientific domains. Recently, Reinforcement Learning with Verifiable Rewards (RLVR) ~\cite{guo2025deepseek,yu2025dapo} has emerged as the dominant paradigm for further amplifying these reasoning capabilities, enabling models to transcend the limitations of supervised fine-tuning. 

Despite these capabilities, the training process of RL remains notoriously complex and resource-intensive~\cite{ouyang2022training}, necessitating the development of more efficient training methods. A key distinction contributing to this inefficiency is the nature of supervision: unlike Supervised Fine-Tuning (SFT), which benefits from dense knowledge transfer via teacher-forcing, RL (specifically RLVR) relies on sparse supervision, typically manifesting as a 1-bit reward signal~\cite{uesato2022solving}. Mechanistically, this sparsity leads to updates being confined to small subnets~\cite{frankle2018lottery,mukherjee2025reinforcement} or sparse parameters~\cite{zhu2025path}, implying significant parameter redundancy during full-parameter RL training. Consequently, there is substantial scope for optimizing RL through parameter-efficient approaches. Recent works~\cite{wang2025tina} have demonstrated that Low-Rank Adaptation (LoRA)~\cite{hulora}—which decomposes weight updates into low-rank matrices to reduce computational cost—can yield competitive performance compared to full-parameter training.

\paragraph{Research Question.} While a proliferation of LoRA variants and PEFT methods has emerged, the application of these techniques in reinforcement learning remains predominantly confined to standard LoRA. This predominance raises a critical uncertainty regarding whether the standard LoRA architecture truly represents the optimal strategy for the distinct optimization dynamics of RL, considering there are many other PEFT variants \textit{e.g.,} DoRA~\cite{liu2024dora} that have been verified to be stronger than LoRA under the fine-tuning scenarios. This anchors our primary research question:

\vspace{-15pt} \begin{center} \begin{researchquestion} Which Parameter-Efficient method is best suited for Reinforcement Learning? \end{researchquestion} \end{center} \vspace{-15pt}

\begin{figure}[t]
    \centering
    \includegraphics[width=0.99\linewidth]{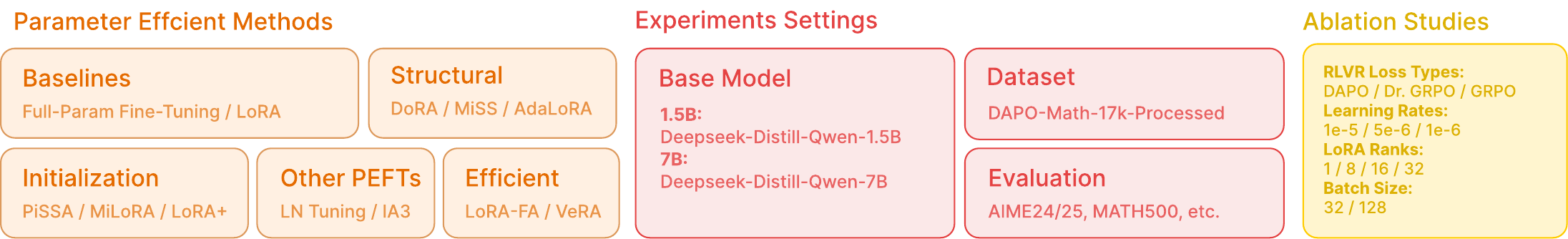}
    \caption{Overview of our evaluation. \textbf{Left}: We systematically evaluate a wide range of parameter-efficient methods categorized. \textbf{Center}: The experimental setup spans diverse base models and is validated on mathematical reasoning datasets. \textbf{Right}: Comprehensive ablation studies are conducted across RLVR loss types, learning rates, LoRA ranks, and batch sizes to ensure the robustness of our findings.}
    \vspace{-0.3cm}
    \label{fig:overview}
\end{figure}

To address this, we conduct the first large-scale, comprehensive evaluation of PEFT methods within RL. Through a multidimensional analysis, we derive actionable insights to guide the community in navigating the development of Parameter-Efficient Reinforcement Learning.

 \paragraph{Experimental settings.} To rigorously investigate these dynamics, we construct a large-scale benchmark using the \texttt{DeepSeek-R1-Distill}~\cite{deepseekai2025deepseekr1incentivizingreasoningcapability} 
 model families. Our experiments span mathematical reasoning tasks including MATH-500~\cite{lightman2023lets}, AIME~\cite{aime24,aime25}, AMC~\cite{numina_math_datasets}, \textit{etc.,} utilizing the RLVR framework on TRL~\cite{vonwerra2022trl}. We evaluate over 12 PEFT variants categorized into structural, initialization-based, and efficiency-driven methods. All methods are tested under controlled conditions with unified hyperparameters (\textit{e.g.,} learning rate, batch size, and rank) to ensure a fair comparison across the distinct optimization landscapes of each adapter.

\paragraph{Key Findings.} Our large-scale empirical analysis challenges the default adoption of standard LoRA, highlighted by three core insights: 
(1) \textbf{Structural variants surpass standard LoRA.} We find that standard LoRA is suboptimal for RLVR. Structural variants \textit{e.g.,} DoRA~\cite{liu2024dora}, MiSS~\cite{kang2025missrevisitingtradeofflora}, and AdaLoRA~\cite{zhang2023adalora} consistently yield superior reasoning accuracy, with DoRA notably outperforming even full-parameter fine-tuning. 
(2) \textbf{SVD-based initialization suffers from spectral misalignment.} Strategies prioritizing principal components \textit{e.g.,} PiSSA~\cite{meng2024pissa} experience training collapse, which we attribute to a fundamental conflict with RL's intrinsic \textit{off-principal} update dynamics. Conversely, initialization methods based on learning rate adjustment \textit{e.g.,} LoRA+~\cite{hayou2024loraplus} prove highly robust. 
(3) \textbf{Less is not always more for Parameter-Efficient RLVR.} Our findings reveals that, while RLVR can tolerate moderate parameter reduction, \textit{e.g.,} freezing half of the weights in LoRA-FA, it exhibits a strict lower bound on expressivity. Extreme compression schemes, such as VeRA~\cite{kopiczko2023vera}, Rank-1 adapters, or exclusive LayerNorm tuning, act as a structural bottleneck that causes performance to collapse, failing to sustain the acquisition of complex reasoning behaviors.

\paragraph{Ablation and Scaling.} Extensive ablations across batch sizes, ranks, and learning rates further substantiate the robustness of our conclusions. Crucially, we extend our validation by scaling to the larger \texttt{DeepSeek-R1-Distill-Qwen-7B} model. The consistent superiority of structural variants across these varying parameter scales confirms that our insights are intrinsic to the RLVR optimization landscape and hold firm regardless of model capacity.

\paragraph{Contributions.}
To the best of our knowledge, this work represents the first systematic study bridging the gap between diverse PEFT methodologies and the specific optimization dynamics of Reinforcement Learning with Verifiable Rewards. Our contributions are summarized as follows:

\begin{itemize}[leftmargin=*, topsep=2pt, itemsep=4pt]
    \item \textbf{\sffamily First Comprehensive PEFT-RLVR Benchmark} (Section~\ref{sec:pre}). We establish a large-scale benchmark evaluating over 12 parameter-efficient methods. We demonstrate that the prevailing practice of defaulting to standard LoRA is suboptimal for RLVR.

    \item \textbf{\sffamily Superiority of Structural Variants} (Section~\ref{sec:results}). We empirically demonstrate that structural variants consistently outperform standard LoRA and frequently surpass full-parameter fine-tuning.

    \item \textbf{\sffamily Mechanism of SVD-based Initialization Failure} (Section~\ref{sec:results}). We uncover a critical failure mode in SVD-informed initialization strategies. Through spectral analysis, we provide a mechanistic explanation: these methods enforce updates on principal components, creating a fundamental structural misalignment with RLVR's intrinsic tendency to operate in the off-principal regime.

    \item \textbf{\sffamily Identification of the Expressivity Floor} (Section~\ref{sec:results}). We identify a distinct performance boundary in parameter efficiency. Our results reveal that extreme parameter reduction methods create an information bottleneck that severely limits the plasticity required for reasoning.
    
    \item \textbf{\sffamily Scalability and Robustness} (Section~\ref{sec:ablation} and~\ref{sec:scale}). We validate the generalizability of our findings by scaling experiments to the 7B parameter regime and conducting extensive ablations on batch sizes, learning rates, and ranks.
\end{itemize}

\section{Preliminaries \& Setup}
\label{sec:pre}

\subsection{Reinforcement Learning with Verifiable Rewards (RLVR)}

\paragraph{Group Relative Policy Optimization (GRPO).} Reinforcement Learning with Verifiable Rewards (RLVR) has emerged as a powerful paradigm for enhancing LLM reasoning by utilizing deterministic verifiers to provide sparse but accurate binary rewards \cite{yu2025dapo,shao2024deepseekmath}. Unlike traditional RLHF, RLVR leverages rule-based feedback (e.g., math correctness or code execution) to elicit complex behaviors such as self-correction and iterative refinement~\cite{shao2024deepseekmath, liu2025understanding}. The foundational framework for many recent advancements is Group Relative Policy Optimization (GRPO), which eliminates the need for a separate critic model by estimating advantages through group statistics \cite{shao2024deepseekmath}. For a given prompt $q$, GRPO samples a group of $G$ responses $\{o_1, \dots, o_G\}$ and optimizes the following surrogate objective:

{
\small
\begin{equation}
    \mathcal{J}_{GRPO}(\theta) = \mathbb{E}_{q \sim \mathcal{D}, \{o_i\} \sim \pi_{\theta_{old}}} \left[ \frac{1}{G} \sum_{i=1}^G \frac{1}{|o_i|} \sum_{t=1}^{|o_i|} \min \left( \frac{\pi_{\theta}(o_{i,t}|q, o_{i,<t})}{\pi_{\theta_{old}}(o_{i,t}|q, o_{i,<t})} \hat{A}_i, \mathrm{clip} \left( \frac{\pi_{\theta}(o_{i,t}|q, o_{i,<t})}{\pi_{\theta_{old}}(o_{i,t}|q, o_{i,<t})}, 1 \pm \epsilon \right) \hat{A}_i \right) \right]
\end{equation}
}

where $\hat{A}_i = \frac{(R_i - \mathrm{mean}(\{R_j\})) }{\mathrm{std}(\{R_j\})}$ represents the standardized advantage within the group \cite{shao2024deepseekmath}.

\paragraph{GRPO Variants and Improvements.} To address challenges such as entropy collapse and training instability in long CoT scenarios, several optimized variants have been proposed. Decoupled Clip and Dynamic sampling Policy Optimization (DAPO) introduces a \textit{Clip-Higher} strategy, which decouples the clipping range into $\epsilon_{low}$ and $\epsilon_{high}$ \cite{yu2025dapo}. By setting a larger $\epsilon_{high}$ \textit{e.g.,} 0.28, DAPO provides more room for low-probability exploration tokens to be uplifted, effectively maintaining policy diversity \cite{yu2025dapo}. Furthermore, DAPO employs \textit{Dynamic Sampling} to filter out prompts where all outputs yield identical rewards \textit{e.g.,} all $0$ or all $1$, ensuring consistent gradient signals and improved sample efficiency~\cite{yu2025dapo}. Another significant refinement is Dr. GRPO, which identifies and mitigates systematic biases inherent in the original GRPO formulation \cite{liu2025understanding}. Dr. GRPO removes the per-response length normalization term $\frac{1}{|o_i|}$, which inadvertently rewards longer incorrect responses while penalizing concise correct ones. Additionally, Dr. GRPO eliminates the group-level standard deviation in advantage estimation to avoid difficulty bias, where questions with low reward variance (too easy or too hard) receive disproportionately high weights~\cite{liu2025understanding}. 

Following the previous works~\cite{shao2024deepseekmath,yu2025dapo,he2025justrl,wang2025tina}, we adopt \textbf{DAPO} as our standard training algorithm and leave other methods as ablation experiments.

\begin{table}[t]
    \centering   
    \small
    \renewcommand{\arraystretch}{1.1}
    \begin{tabular}{lll}
    \toprule
         \textbf{Method}&  \textbf{Forward}&  \textbf{Initialization} \\
         \midrule
         \textit{\textbf{Baseline}}& &\\
         \quad Full-Param Fine-Tuning& $y = \boldsymbol{W}_{0} x$& N/A\\
         \quad LoRA~\cite{hulora} & $y = \boldsymbol{W}_{0} x + \frac{\alpha}{r}\boldsymbol{BA} x$ &$\boldsymbol{A} \sim N(0, \sigma^2)$, $\boldsymbol{B} \sim 0$ \\
         \textit{\textbf{Structural}}& &\\
         \quad DoRA~\cite{liu2024dora}& $y = \boldsymbol{m}(\ \boldsymbol{W}_{0}x+\boldsymbol{BA}x ~ / ~ {\|\boldsymbol{W}_{0}+\boldsymbol{BA}\|_{c}})$&$\boldsymbol{A} \sim \mathrm{Rect. KaimingUnif}, \ \boldsymbol{B} \sim 0$\\
         \quad MiSS~\cite{kang2025missrevisitingtradeofflora}& $y = \boldsymbol{W_0} x + \operatorname{expand}(\boldsymbol{D}) x$&$\boldsymbol{D} \sim 0$  \\
         \quad AdaLoRA~\cite{zhang2023adalora}& $y = \boldsymbol{W}_{0} x + \boldsymbol{P\Lambda Q} x$&$\boldsymbol{\Lambda} \sim 0, \ \boldsymbol{P}, \boldsymbol{Q} \sim N(0, \sigma^2)$ \\
         \textit{\textbf{Initialization}}& &\\
         \quad PiSSA~\cite{meng2024pissa}&$y = (\boldsymbol{W}_{0} - \boldsymbol{BA}) x + \boldsymbol{BA} x$& $\boldsymbol{A} = U_{[:,:r]}S_{[:r,:r]}^{1/2},\ \boldsymbol{B} = S_{[:r,:r]}^{1/2}V_{[:,:r]}^{\top}$  \\
         \quad MiLoRA~\cite{wang2025milora}& $y = (\boldsymbol{W}_{0} -\boldsymbol{BA})x + \boldsymbol{BA} x$&$\boldsymbol{A} = U_{[:,r:]}S_{[r:,r:]}^{1/2},\ \boldsymbol{B} = S_{[r:,r:]}^{1/2}V_{[:,r:]}^{\top}$\\
         \quad LoRA+~\cite{hayou2024loraplus}& $y = \boldsymbol{W}_{0} x + \boldsymbol{BA} x$ &$\eta_{B} = \lambda \eta_{A}$ (Ratio of Learning Rates)\\
         \quad rsLoRA~\cite{kalajdzievski2023rank} & $y = \boldsymbol{W}_{0} x + \frac{\alpha}{\sqrt{r}}\boldsymbol{BA} x$ & $\boldsymbol{A} \sim N(0, \sigma^2), \boldsymbol{B} \sim 0$ \\
         \textit{\textbf{Efficiency}}& &\\
         \quad LoRA-FA~\cite{zhang2023lora}& $y = \boldsymbol{W}_{0} x + \boldsymbol{BA} x$ & $\boldsymbol{A} \sim N(0, \sigma^2) \text{ (Frozen)}, \boldsymbol{B} \sim 0$ \\
         \quad VeRA~\cite{kopiczko2023vera}& $y = \boldsymbol{W}_{0} x + \Lambda_{b}\boldsymbol{B}\Lambda_{d}\boldsymbol{A}x$ & $\boldsymbol{A}, \boldsymbol{B} \text{ Frozen Random}, \boldsymbol{d} \sim 0.1, \boldsymbol{b} \sim 0$ \\
         \textit{\textbf{Other PEFTs}}&  & \\
         \quad LN Tuning~\cite{qi2022parameterefficienttuninglayernormalization}& $y = \frac{g}{\sigma} \odot (x - \mu) + b$ & Pre-trained $g$ (gain) and $b$ (bias) \\
         \quad IA$^3$~\cite{Liu2022FewShotPF}& $x' = x \odot l$ & $l \sim 1$ (Rescaling vectors for $K, V, \text{FFN}$) \\
    \bottomrule
    \end{tabular}
    \vspace{0.1cm}
    \caption{A variety of PEFT methods are listed, each with its specific update formulation and initialization strategy. LN denotes Layernorm.}
    \vspace{-0.6cm}
    \label{tab:overview}
\end{table}

\subsection{PEFT Methods}
\paragraph{Low-Rank Adaptation (LoRA).} 
LoRA~\cite{hulora} hypothesizes that the change in weights during adaptation has a low intrinsic rank. Given a pre-trained weight matrix $\mathbf{W}_0 \in \mathbb{R}^{d_{\text{out}} \times d_{in}}$, LoRA freezes $\mathbf{W}_0$ and constrains the update $\Delta \mathbf{W}$ by decomposing it into the product of two low-rank matrices $\mathbf{B} \in \mathbb{R}^{d_{\text{out}} \times r}$ and $\mathbf{A} \in \mathbb{R}^{r \times d_{in}}$, where the rank $r \ll \min(d_{in}, d_{out})$. The forward pass is formalized as:
\begin{equation}
    \mathbf{h} = \mathbf{W}_0 \mathbf{x} + \Delta \mathbf{W} \mathbf{x} = \mathbf{W}_0 \mathbf{x} + \frac{\alpha}{r} \mathbf{B}\mathbf{A}\mathbf{x},
\end{equation}
where $\alpha$ is a constant scaling factor. In the standard implementation, $\mathbf{A}$ is initialized with random Gaussian noise, while $\mathbf{B}$ is initialized to zero, ensuring that $\Delta \mathbf{W} = 0$ at the beginning of training.

\paragraph{PEFT Methods Selection.}
While LoRA remains the most prominent approach, the landscape of parameter-efficient methods has expanded significantly to encompass a diverse array of alternative methodologies. To provide a thorough evaluation, we categorize these adopted PEFT methods into five distinct groups based on their design paradigms, as illustrated in Figure~\ref{fig:overview} and Table~\ref{tab:overview}:

\begin{itemize}[leftmargin=*, topsep=2pt, itemsep=4pt]
    \item \textbf{\sffamily Baselines}: We employ Full-Parameter Fine-Tuning and standard LoRA~\cite{hulora} as our primary benchmarks to establish the performance upper bound and the standard efficiency baseline ($y = \boldsymbol{W}_{0} x + \boldsymbol{BA} x$), respectively.
    
    \item \textbf{\sffamily Structural Variants}: This category encompasses methods that \textit{fundamentally alter the architectural formulation} beyond the standard additive product $\boldsymbol{BA}$. Instead of the fixed low-rank decomposition, these methods introduce novel structural components. We include \textbf{DoRA}~\cite{liu2024dora}, which decouples magnitude and direction ($y = \boldsymbol{m} \frac{\boldsymbol{W}x}{||\boldsymbol{W}||}$); \textbf{AdaLoRA}~\cite{zhang2023adalora}, which employs an SVD-like adaptive rank structure ($y = \boldsymbol{W}_{0} x + \boldsymbol{P\Lambda Q} x$); and others like \textbf{MiSS}~\cite{kang2025missrevisitingtradeofflora} 
    that utilize distinct sub-network selection.

    \item \textbf{\sffamily Initialization Strategies}\footnote{In this context, we define initialization-based methods broadly to include all strategies that are configured at the onset of training without altering the standard LoRA architecture ($\boldsymbol{W}_{0} + \boldsymbol{BA}$), encompassing both weight initialization and the pre-specification of optimization dynamics.}: These methods \textit{retain the standard adapter architecture} but intervene in the initialization state or optimization dynamics to accelerate convergence. We evaluate signal-informed strategies like \textbf{PiSSA}~\cite{meng2024pissa} and \textbf{MiLoRA}~\cite{wang2025milora}, which initialize matrices $\boldsymbol{A}$ and $\boldsymbol{B}$ using the Principal Components (SVD) of $\boldsymbol{W}_0$ rather than random Gaussian noise. We also examine methods that adjust the training dynamics, such as \textbf{LoRA+}~\cite{hayou2024loraplus}, which uses differentiated learning rates ($\eta_B \gg \eta_A$), and \textbf{rsLoRA}~\cite{kalajdzievski2023rank}, which employs stable rank scaling factors.
    
    \item \textbf{\sffamily Efficiency-Oriented Variants}: Driven by hardware constraints, this category investigates methods designed to minimize memory footprints. We evaluate \textbf{LoRA-FA}~\cite{zhang2023lora} (freezing $\boldsymbol{A}$) and \textbf{VeRA}~\cite{kopiczko2023vera} (freezing random projection matrices and training only scaling vectors).
    
    \item \textbf{\sffamily Other PEFT Mechanisms}: Finally, we assess approaches that diverge from the weight-update paradigm entirely. This includes \textbf{IA$^3$}~\cite{liu2022few}, which scales activation vectors via element-wise multiplication, and \textbf{LayerNorm Tuning}~\cite{qi2022parameterefficienttuninglayernormalization}, to evaluate the efficacy of alternative adaptation mechanisms.
\end{itemize}

\paragraph{PEFT Settings.} We benchmark standard LoRA and other PEFT methods. It is worth noting that following~\citet{schulman2025lora} and~\citet{wang2025tina}, we target all linear modules ($\{q, k, v, o, gate, up, down\}\_proj$), as this configuration has been demonstrated to yield superior performance in prior studies~\cite{schulman2025lora}. We set rank 32, dropout rate 0.05, and alpha 64 for all PEFT methods.





\subsection{Models and Datasets}
\paragraph{Base Models.} Our selection of base models is guided by three key criteria: (1) Following the standard RL paradigm, we select models that have undergone SFT as a cold start phase to ensure sufficient initial reasoning capabilities and reasoning format; (2) We employ models across different parameter scales to disentangle size-specific effects. Based on these principles, we utilize two reasoning models: \textbf{DeepSeek-R1-Distill-Qwen-1.5B}~\cite{deepseekai2025deepseekr1incentivizingreasoningcapability},
and \textbf{DeepSeek-R1-Distill-Qwen-7B}~\cite{deepseekai2025deepseekr1incentivizingreasoningcapability}. 

\paragraph{Training Dataset.} We utilize the \texttt{open-r1/DAPO-Math-17k-Processed} dataset~\cite{yu2025dapo}, which comprises approximately 17.4k high-quality mathematical queries and has been validated by prior research. To enforce structured reasoning, we impose a strict output format, requiring the model to enclose reasoning traces within \texttt{<think>...</think>} tags and encapsulate the final answer using \texttt{\textbackslash\textbackslash boxed\{\}}.



\subsection{Training Settings}

\paragraph{Reward Mechanism.} We employ a strict outcome-based reward. The generated answers are extracted and verified against ground truth using a combination of \texttt{latex2sympy} and \texttt{math\_verify}. The reward is binary: $R=1$ for mathematically equivalent answers and $R=0$ otherwise. The overall reward recipe follows the principles of JustRL \cite{he2025justrl}.

\paragraph{Hyperparameter and Other Details.} 
We utilize \texttt{Accelerate} with DeepSpeed ZeRO-2 optimization (offloading optimizer states) to minimize memory usage. For rollout generation, we employ the \texttt{vLLM} engine in co-location mode to maximize throughput. Following the previous settings of LoRA RL training~\cite{wang2025tina}, we generate $G=8$ rollouts per prompt and use a constant learning rate of $1\text{e-}5$ with no warmup. The training is conducted with a maximum prompt length of 512 and a completion length of $16384$ tokens. For the RLVR objective, we set the DAPO epsilon to $0.28$ \textit{i.e.,} \textit{clip-higher} and do not employ a KL coefficient ($\beta$). Regarding the batch configurations, the 1.5B model is trained with a per-device batch size of 4 and a global batch size of 128 over 1,024 steps; for the 7B model, we use a per-device batch size of 1 and a global batch size of 32 over 8,192 steps. In both settings, the gradient accumulation steps are fixed at 8. We do not apply complicated strategies \textit{e.g.,} multi-stage training, as it has been proven that simple RL scaling~\cite{he2025justrl} can still achieve competitive results.

\subsection{Evaluation}
\paragraph{Benchmark Selections.}
We evaluate models using the mathematics suite. following the benchmark selection in Table~\ref{tab:bench} from previous work including~\citet{he2025justrl}, \citet{guo2025deepseek} and \citet{wang2025tina}. The benchmarks include MATH-500~\cite{lightman2023lets}, AMC23~\cite{numina_math_datasets}, AIME24/25~\cite{aime24,aime25}, Minerva~\cite{lewkowycz2022solving}, and HMMT~\cite{balunovic2025matharena}.

\begin{wraptable}{r}{0.55\textwidth}
    \centering
    \small
    \vspace{-0.4cm}
    \begin{tabular}{lll}
        \toprule
        Benchmark& Nums& Eval Method\\
        \midrule
        AIME24~\cite{aime25}& 30& $\mathrm{Avg}@32$\\
        AIME25~\cite{aime25}& 30& $\mathrm{Avg}@32$\\
        MATH500~\cite{lightman2023lets}& 500& $\mathrm{Avg}@4$\\
        Minerva~\cite{lewkowycz2022solving}& 272 & $\mathrm{Avg}@4$\\
        AMC~\cite{numina_math_datasets}& 40 & $\mathrm{Avg}@32$\\
        HMMT~\cite{balunovic2025matharena}& 30 & $\mathrm{Avg}@32$\\
        \bottomrule
    \end{tabular}
    \caption{Overview of the mathematical reasoning datasets used in our study, including the number of test samples and the specific evaluation metrics (\textit{e.g.,} $Avg@k$) employed for each benchmark.}
    \label{tab:bench}
    \vspace{-0.3cm}
\end{wraptable}

\paragraph{Evaluation Settings.} For evaluation generation, we use a temperature of $0.6$ and top-$p$ of $0.95$ to allow for diverse reasoning paths, with a maximum token limit of 32768 to accommodate long chain-of-thought processes. The random seed is fixed at 42 for reproducibility. Considering that standard benchmarks, such as AIME, contain a relatively small number of questions, the statistical variation of the evaluation results can be significantly influenced. To mitigate this issue and enhance the robustness of our metrics, we compute the $18.0$ $\mathrm{Avg}@k$ for each problem, defined as the average accuracy across $k$ generations. We also evaluate $\mathrm{Pass}@1$ in $k$ \textit{i.e.,} if there is on correct answer in $k$ generations, we consider this problem as solved.

\section{Results and Analysis}

\subsection{Main Results}
\label{sec:results}
\paragraph{LoRA is Definitely not the optimal choice for RL.} A salient observation from our experiments is the consistent superiority of some LoRA-variants \textit{e.g.,} DoRA and MiSS.

\begin{table}[t]
    \centering
    \setlength{\tabcolsep}{4pt}
    \small
    \begin{tabular}{lccccccccccccc}
    \toprule
    \textbf{Methods} & \textbf{Avg.} & \multicolumn{2}{c}{\textbf{AIME24@32}}& \multicolumn{2}{c}{\textbf{AIME25@32}}& \multicolumn{2}{c}{\textbf{AMC@32}}& \multicolumn{2}{c}{\textbf{HMMT@32}}& \multicolumn{2}{c}{\textbf{MATH500@4}}& \multicolumn{2}{c}{\textbf{Minerva@4}}\\
     & &  Avg.&Pass&  Avg.&Pass&  Avg.&Pass&  Avg.&Pass&  Avg.&Pass& Avg.&Pass\\
    \midrule
    \textit{\textbf{Baseline}} & &  &&  &&  &&  &&  &&  &\\
 Base& $40.5$& $32.4$& $76.7$& $22.2$& $33.3$& $69.0$& $90.0$& $9.9$& $23.3$& $72.0$& $86.8$& $14.2$&$25.0$\\
    Full & $44.9$ & $34.9$ & $56.7$ & $23.8$ & $46.7$ & $68.8$ & $92.5$ & $13.5$ & $40.0$ & $74.8$ & $88.6$ & $17.0$ & $26.5$ \\
    LoRA & $42.5$ & $33.2$ & $60.0$ & $22.9$ & $36.7$ & $64.4$ & $95.0$ & $13.3$ & $33.3$ & $72.1$ & $87.4$ & $13.5$ & $23.5$ \\
    \textit{\textbf{Structural}} & &  &&  &&  &&  &&  &&  &\\
    \rowcolor{red!10} AdaLoRA & $44.2$ & $29.4$ & $53.3$& $26.7$ & $50.0$& $68.9$ & $95.0$& $14.1$ & $40.0$& $73.8$ & $88.0$& $15.5$ & $27.5$\\
    \rowcolor{red!10} DoRA & $46.6$ & $39.0$ & $80.0$ & $28.8$ & $43.3$ & $71.9$ & $95.0$ & $13.4$ & $36.7$ & $75.8$ & $90.0$ & $14.7$ & $26.5$ \\
    \rowcolor{red!10} MiSS & $43.4$ & $27.5$ & $50.0$ & $23.3$ & $26.7$ & $68.6$ & $95.0$ & $15.8$ & $33.3$ & $72.4$ & $90.4$ & $16.5$ & $30.1$ \\
    \textit{\textbf{Efficient}} & &  &&  &&  &&  &&  &&  &\\
    \rowcolor{yellow!10} LoRA-FA & $43.0$&  $29.2$&$53.3$&  $22.7$&$46.7$&  $65.0$&$95.0$ &  $15.1$&$36.7$ &  $73.5$&$87.2$&  $15.6$&$26.8$\\
    \rowcolor{yellow!10} VeRA & $40.7$ & $29.1$ & $60.0$ & $21.7$ & $36.7$ & $61.5$ & $95.0$ & $14.4$ & $30.0$ & $69.7$ & $85.6$ & $13.1$ & $24.6$ \\
    \textit{\textbf{Initialization}} & &  &&  &&  &&  &&  &&  &\\
    \rowcolor{orange!10} LoRA+ & $43.9$ & $28.1$ & $50.0$ & $25.9$ & $50.0$& $70.0$ & $95.0$ & $16.5$ & $36.7$ & $72.2$ & $90.4$ & $15.4$ & $26.5$ \\
    \rowcolor{orange!10} rsLoRA & $42.3$ & $29.2$ & $43.3$ & $23.3$ & $30.0$ & $63.4$ & $92.5$ & $14.4$ & $36.7$ & $72.6$ & $87.6$ & $14.4$ & $27.2$ \\
    \rowcolor{orange!10} MiLoRA & $18.0$ & $4.2$ & $6.7$ & $0.0$ & $0.0$ & $19.6$ & $47.5$ & $0.0$ & $0.0$ & $44.5$ & $63.4$ & $11.7$ & $19.9$ \\
    \rowcolor{orange!10} PiSSA & $0.2$ & $0.0$ & $0.0$ & $0.0$ & $0.0$ & $0.0$ & $0.0$ & $0.0$ & $0.0$ & $0.6$ & $1.0$ & $0.1$ & $0.4$ \\
    \textit{\textbf{Other PEFTs}} & &  &&  &&  &&  &&  &&  &\\
    \rowcolor{gray!10} IA$^3$ & $22.3$ & $0.7$ & $6.7$ & $3.5$ & $10.0$ & $25.2$ & $57.5$ & $0.6$ & $3.3$ & $55.4$ & $71.4$ & $12.8$ & $22.1$ \\
    \rowcolor{gray!10} LN Tuning & $41.8$ & $31.7$ & $56.7$ & $23.8$ & $26.7$ & $65.1$ & $95.0$ & $11.8$ & $23.3$ & $69.9$ & $85.0$ & $14.1$ & $25.0$ \\
    \bottomrule
    \end{tabular}
    \vspace{0.1cm}
    \caption{Comparison of accuracy and pass scores. All values are reported in percentages.}
    \label{tab:main_results}
    \vspace{-0.4cm}
\end{table}

\begin{centering}
    \begin{findings}
        \textbf{\sffamily Finding 1:} Standard LoRA is \textbf{\sffamily suboptimal} for RLVR. \textbf{\sffamily Structural variants}, that decouple learning dynamics (DoRA), sharding parameters (MiSS) or allocate parameters adaptively (AdaLoRA) , currently represent the optimal parameter-efficient choices for RLVR beyond LoRA. {\bfseries\sffamily So stop using LoRA for RLVR training!}
    \end{findings}
\end{centering}

While standard LoRA ($42.5\%$) serves as a respectable baseline, it consistently trails behind full-parameter fine-tuning ($44.9\%$), suggesting a limitation in its rigid low-rank constraint when facing the complex policy shifts required by RL. In contrast, structural variants effectively bridge or even exceed this gap. DoRA breaks the ceiling with an overall average of $46.6\%$, surpassing the full-parameter baseline across multiple benchmarks (\textit{e.g.,} AIME and AMC). Similarly, AdaLoRA ($44.2\%$) and MiSS ($43.4\%$) consistently outperform standard LoRA. We attribute this superiority to the mitigation of the optimization rigidity inherent in standard LoRA, and we hypothesize that this stems from a fundamental alignment between the architectural inductive biases of these variants and the unique optimization dynamics of RLVR.

\paragraph{Less is not always more for parameter-efficient RLVR.} While recent findings suggest RLVR is compatible with low-rank updates~\citep{schulman2025lora}, our results identify a critical expressivity floor. We observe that while moderate efficiency gains are sustainable, extreme parameter reduction methods fail to capture the complex policy shifts required for reasoning. As detailed in Table~\ref{tab:train_p_acc}, there exists a clear boundary in performance based on the adaptation mechanism. Methods that retain low-rank matrix structures, such as LoRA-FA (which freezes projection matrices $\mathbf{A}$ and trains only $\mathbf{B}$), maintain competitive performance comparable to standard LoRA. This indicates that the RLVR signal, though sparse, is sufficient to drive updates in a semi-frozen low-rank subspace.

\begin{centering}
    \begin{findings}
        \textbf{\sffamily Finding 2:} \textbf{\sffamily RLVR demands a minimum threshold of expressivity.} While moderate reduction (\textit{e.g.,} LoRA-FA) is effective, extreme parameter reduction methods (\textit{e.g.,} VeRA, IA$^3$) that rely on vector-only updates lack the necessary plasticity to reorient reasoning circuits. 
    \end{findings}
\end{centering}

\begin{wraptable}{r}{0.45\textwidth}
    \centering
    \small
    \vspace{-0.0cm}
    \begin{tabular}{lcc}
        \toprule
        Method& Train Param.& Overall Acc. (\%)\\
        \midrule
        Full& $100$\%& $\mathbf{44.9}$\\
        LoRA& $1.55\%$& $42.5$\\
        LoRA Rank 1& $0.0015\%$& $40.5$\\
        \rowcolor{orange!10} MiSS& $0.99\%$& $\underline{43.4}$\\
        \rowcolor{yellow!10} VeRA& $0.0029\%$& $40.7$\\
        \rowcolor{yellow!10} LN Tuning& $0.0035\%$& $41.8$\\
        \bottomrule
    \end{tabular}
    \caption{Trainable parameters and overall accuracy of full-parameter fine-tuning, standard LoRA, MiSS, VeRA and LN Tuning.}
    \label{tab:train_p_acc}
    \vspace{-0.1cm}
\end{wraptable}

While RLVR can tolerate moderate parameter reduction, we find it fails under \textit{extreme parameter reduction}. For instance, VeRA—which freezes both low-rank matrices and learns only scaling vectors—drops to 40.7\% accuracy, and IA$^3$ suffers a severe degradation to 22.3\%. These results indicate that RLVR requires a minimum threshold of trainable adapter capacity to succeed. Unlike supervised fine-tuning, the optimization process in RLVR appears to demand higher expressivity in the trainable space; reducing this space to mere scaling vectors (as in VeRA, IA$^3$, or LN-tuning) creates a bottleneck that prevents the model from effectively learning complex reasoning behaviors.


\paragraph{SVD-based Initialization Misaligns with RL Optimization.} 

\begin{figure}[ht]
    \centering
    \vspace{-0.1cm}
    \begin{minipage}[c]{0.73\textwidth}
        \centering
        \includegraphics[width=\linewidth]{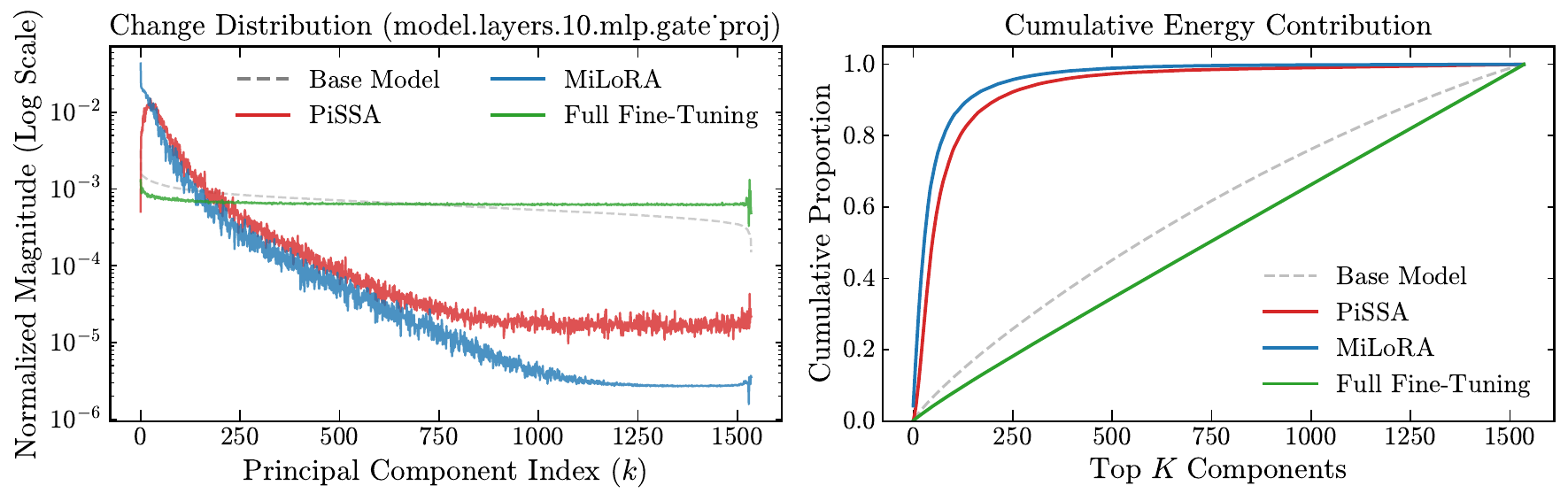}
    \end{minipage}
    \begin{minipage}[c]{0.23\textwidth}
        \centering
        \includegraphics[width=\linewidth]{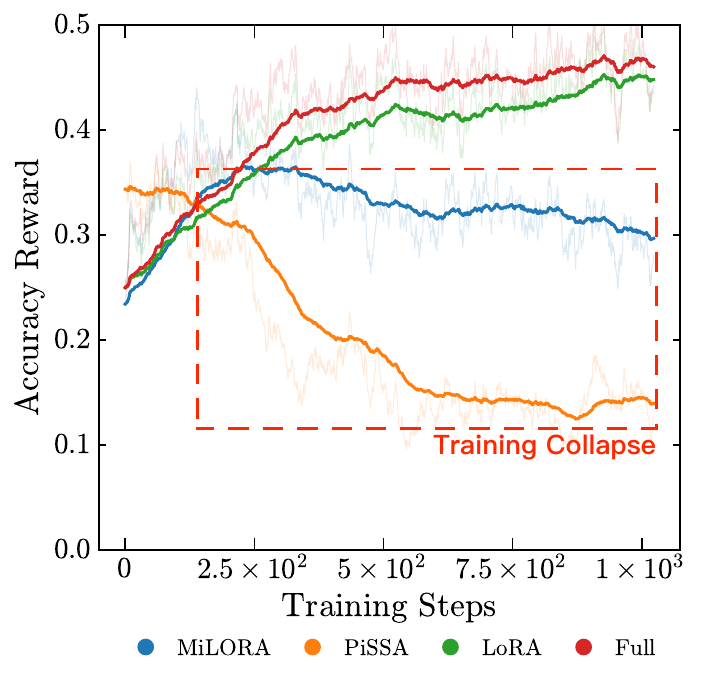}
    \end{minipage}
    \caption{\textbf{Left:} Normalized magnitude of updates across singular value indices. \textbf{Center}: Cumulative proportion of energy explained by the top-$k$ components. \textbf{Right:} Accuracy reward curves during training, illustrating the performance collapse of SVD-based initializations in the RLVR setting compared to standard baselines.}
    \label{fig:svd}
    \vspace{-0.2cm}
\end{figure}

The substantial underperformance of initialization strategies derived from Singular Value Decomposition warrants a mechanistic explanation. As shown in Table~\ref{tab:main_results}, PiSSA suffers a catastrophic collapse to near-zero accuracy (0.2\%), while MiLoRA (18.0\%) significantly trails standard baselines. Recent work by~\citet{zhu2025path} reveals that RLVR operates in an \textit{off-principal regime}: unlike SFT, which targets high-magnitude principal weights, RLVR updates consistently localize to low-curvature, non-principal subspaces to preserve the pre-trained spectral geometry. Based on this characteristic, PiSSA fails predictably: by explicitly restricting updates to the principal subspace ($U_{[:r]}, V_{[:r]}$), it imposes a structural bias that directly conflicts with the intrinsic requirement of RLVR, leading to the observed collapse (0.2\% accuracy).

A more critical finding, however, contradicts the intuitive extension of the off-principal theory. Theoretically, MiLoRA initializes adapters using minor singular components ($U_{[r:]}, V_{[r:]}$), which presumably aligns with the off-principal nature of RLVR. Yet, our empirical results (see Figure~\ref{fig:svd}, Right) show it first trains with a well reward increasing and then fails to converge (18.0\%). 
We analyze this failure through the lens of spectral analysis on weight updates. Our Full Fine-Tuning results (Figure~\ref{fig:svd}, Left, Green line) exhibit a uniform distribution of updates across the entire singular value spectrum, and Figure~\ref{fig:svd} (Left, Blue line) uncovers the mechanism: despite being initialized in the off-principal subspace, the final updates of MiLoRA exhibit a sharp spike at the dominant principal components ($k \approx 0$), behaving almost identically to PiSSA.

Despite the theoretical alignment, MiLoRA fails due to the disparity between initialization and gradient flow. We formalize the update dynamics at step $t$ as:
\begin{equation}
    \Delta \boldsymbol{W}_{t+1} \leftarrow \Delta \boldsymbol{W}_t - \eta \nabla \mathcal{L}(\boldsymbol{W}_t), \ \|\Delta \boldsymbol{W}_{0}\|_F = \|\boldsymbol{B}_{init}\boldsymbol{A}_{init}\|_F \approx 0 \ \mathrm{when} \ t=0.
\end{equation}
The minor singular values used for initialization satisfy $\sigma_{tail} \to 0$. Consequently, the initial adapter state effectively collapses to zero. This renders the intended structural constraint numerically non-existent. Without a significant initial bias $\|\Delta \boldsymbol{W}_{0}\|_F$, the optimization trajectory is dictated by the spectral properties of the gradient $\nabla \mathcal{L}$. Since the gradient aligns with the directions of maximum variance (principal components $\boldsymbol{U}_{:k}$), the update projects onto the principal subspace where $    \langle \nabla \mathcal{L}, \boldsymbol{U}_{:k} \rangle \gg \langle \nabla \mathcal{L}, \boldsymbol{U}_{k:} \rangle$. Consequently, the updates are immediately reoriented by the dominant principal gradients, causing the model to degenerate from the off-principal regime back to the principal subspace, as evidenced by the spectral spike in Figure~\ref{fig:svd}.

 \begin{centering}
    \begin{findings}
        \textbf{\sffamily Finding 3:} Existing SVD-based initialization strategies are unsuitable for RLVR. PiSSA fails due to a design conflict (forcing principal updates), creating a structural mismatch with RLVR's intrinsic tendency to \textbf{\sffamily learn off the principals}~\cite{zhu2025path}, resulting in training collapse. However, we find that MiLoRA fails due to optimization instability—negligible initialization magnitude causes the model to \textbf{\sffamily degenerate back into principal-component updates}, failing to maintain the necessary off-principal trajectory.
    \end{findings}
\end{centering}

\begin{wrapfigure}{r}{0.47\textwidth}
    \vspace{-1cm}
    \centering
    \small
    \begin{tabular}{lc}
        \toprule
        \textbf{Ablation Term}& \textbf{Settings}\\
        \midrule
        Batch Size& 32, 128\\
        RLVR Algorithm& GRPO, DAPO, Dr. GRPO\\
        Learning Rate& $1 \times 10^{-5}$, $5 \times 10^{-6}$, $1 \times 10^{-6}$\\
        Rank& 1, 8, 16, 32\\

        \bottomrule
    \end{tabular}
    \caption{Hyperparameters for RLVR training across model scales.}
    \label{tab:ablation_hyperparams}
    \vspace{-0.2cm}
\end{wrapfigure}

\subsection{Ablation Studies}
\label{sec:ablation}

To rigorously validate the robustness of our findings and disentangle the influence of hyperparameter choices from intrinsic method efficacy, we conducted a comprehensive series of ablation studies. As summarized in Table~\ref{tab:ablation_hyperparams}, we systematically varied key training configurations across four orthogonal dimensions: batch size, reinforcement learning algorithms, learning rates, and LoRA ranks.

\begin{wrapfigure}{r}{0.35\textwidth}
    \centering
    \small
    \vspace{-0.5cm}
    \includegraphics[width=0.35\textwidth]{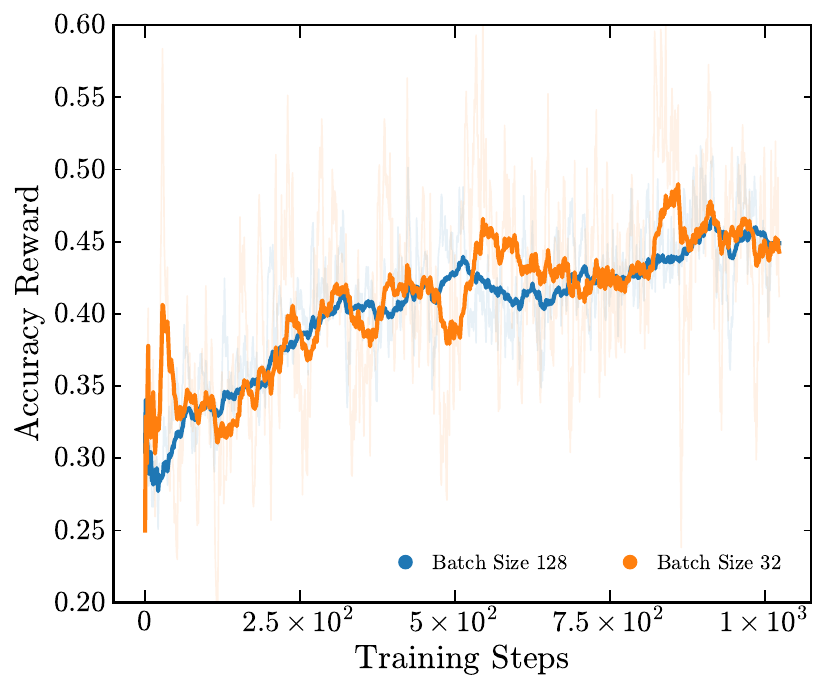}
    \caption{Accuracy reward of batch size 128 and 32.}
    \label{tab:train_bsz_acc}
    \vspace{-0.5cm}
\end{wrapfigure}

\paragraph{RLVR Training Batch Size.} Recent empirical studies in SFT suggest that LoRA efficacy is inversely correlated with batch size, favoring a \textit{small-batch, high-frequency} update regime~\cite{schulman2025lora}. The prevailing hypothesis attributes this to the high information density of SFT's hard teacher-forcing objective, which can saturate the limited capacity of low-rank adapters when processing large batches. We hypothesized that this constraint would be relaxed in RLVR, as the supervision signal consists only of sparse, scalar rewards rather than dense token-level targets. Our results (see Table~\ref{tab:task_avg_comparison_updated} and Figure~\ref{tab:train_bsz_acc}) confirm that: reducing the batch size to 32 yielded an slightly higher average accuracy of 42.5\%. Notably, on the challenging AIME 2024 benchmark, the larger batch size actually outperformed the smaller one. This indicates that the \textit{small batch} heuristic from SFT does transfer to RLVR but perform less well.

\begin{table}[ht]
    \centering
    \setlength{\tabcolsep}{4pt}
    \small
    \begin{tabular}{lccccccccccccc}
    \toprule
    \textbf{Methods} & \textbf{Avg.} & \multicolumn{2}{c}{\textbf{AIME24@32}}& \multicolumn{2}{c}{\textbf{AIME25@32}}& \multicolumn{2}{c}{\textbf{AMC@32}}& \multicolumn{2}{c}{\textbf{HMMT@32}}& \multicolumn{2}{c}{\textbf{MATH500@4}}& \multicolumn{2}{c}{\textbf{Minerva@4}}\\
     & &  Avg.&Pass&  Avg.&Pass&  Avg.&Pass&  Avg.&Pass&  Avg.&Pass& Avg.&Pass\\
    \midrule
    \textit{\textbf{Baseline}} & &  &&  &&  &&  &&  &&  &\\
    Full & $44.9$ & $34.9$ & $56.7$ & $23.8$ & $46.7$ & $68.8$ & $92.5$ & $13.5$ & $40.0$ & $74.8$ & $88.6$ & $17.0$ & $26.5$ \\
    LoRA & $42.5$ & $33.2$ & $60.0$ & $22.9$ & $36.7$ & $64.4$ & $95.0$ & $13.3$ & $33.3$ & $72.1$ & $87.4$ & $13.5$ & $23.5$ \\
     \multicolumn{14}{l}{- \textit{Bsz $128$, learning rate $1 \times 10^{-5}$, DAPO}}\\
    \textit{\textbf{Batch Size}}& &  &&  &&  &&  &&  &&  &\\
    \rowcolor{gray!10} $32$ Bsz& $43.0$& $28.4$& $50.0$& $24.7$& $43.3$& $67.6$& $90.0$& $15.0$& $33.3$& $72.0$& $86.8$& $14.3$& $25.0$\\
    \textit{\textbf{Learning Rate}}& &  &&  &&  &&  &&  &&  &\\
    \rowcolor{orange!10}$1\times 10^{-5}$& $42.3$ & $29.2$ & $43.3$ & $23.3$ & $30.0$ & $63.4$ & $92.5$ & $14.4$ & $36.7$ & $72.6$ & $87.6$ & $14.4$ & $27.2$ \\
    \rowcolor{orange!10}$5\times 10^{-6}$& $42.3$ & $30.4$ & $50.0$ & $18.2$ & $36.7$ & $65.8$ & $92.5$ & $13.9$ & $40.0$ & $73.1$ & $87.2$ & $14.9$ & $27.9$ \\
    \textit{\textbf{Rank}}& &  &&  &&  &&  &&  &&  &\\
    \rowcolor{red!10}$1$ & $40.5$ & $25.7$ & $53.3$ & $21.7$ & $36.7$ & $61.7$ & $95.0$ & $13.4$ & $23.3$ & $70.6$ & $86.4$ & $13.9$ & $24.3$ \\
    \rowcolor{red!10}$8$ & $42.3$ & $28.9$ & $43.3$ & $22.1$ & $40.0$ & $65.4$ & $97.5$ & $15.4$ & $40.0$ & $72.2$ & $87.0$ & $13.5$ & $25.4$ \\
    \rowcolor{red!10}$16$ & $43.9$& $31.2$& $56.7$& $23.8$& $36.7$& $67.3$& $95.0$& $14.7$& $43.3$& $73.8$& $89.6$& $16.1$& $29.4$\\
    \rowcolor{yellow!10} Dr. GRPO& $42.0$ & $28.9$ & $50.0$ & $25.4$ & $43.3$ & $65.6$ & $97.5$ & $13.0$ & $33.3$ & $70.1$ & $87.6$ & $14.5$ & $28.3$ \\
    \rowcolor{yellow!10} GRPO& $40.5$ & $25.2$ & $43.3$ & $15.8$ & $33.3$ & $64.5$ & $97.5$ & $12.7$ & $30.0$ & $71.0$ & $86.8$ & $15.9$ & $28.3$ \\
    \bottomrule
    \end{tabular}
    \vspace{0.1cm}
    \caption{Comparison of accuracy and pass scores. All values are reported in percentages (std dev removed for clarity).}
    \label{tab:task_avg_comparison_updated}
    \vspace{-0.4cm}
\end{table}

\paragraph{RLVR Algorithms.} We further investigated whether the efficacy of parameter-efficient methods is sensitive to the specific formulation of the reinforcement learning objective. We evaluated standard LoRA across three representative RLVR algorithms: GRPO~\cite{shao2024deepseekmath}, DAPO~\cite{yu2025dapo}, and Dr. GRPO~\cite{liu2025understanding}, which employ varying strategies for advantage estimation and regularization. Our experiments reveal a remarkable degree of algorithmic invariance; the performance of LoRA and other PEFT methods remains consistent across these methods, with no statistically significant deviation in reasoning accuracy. This suggests that the effectiveness of parameter efficient methods in this domain is driven by the fundamental dynamics of learning from sparse, verifiable rewards, rather than being contingent on specific loss function nuances (such as the specific implementation of KL penalties or ratio clipping). Consequently, optimal PEFT choices like DoRA are likely transferable across the broader landscape of RLVR algorithms.

\paragraph{Learning Rate.} Our results (see Table~\ref{tab:task_avg_comparison_updated}) corroborate the scaling laws proposed in \citet{schulman2025lora}, confirming that learning rate magnitude is a decisive factor in RLVR stability. We observe that the optimal performance was consistently achieved at scales $\text{LR} = M_{\text{LoRA}} \cdot \left( \frac{2000}{\text{hidden size}} \right)^{\text{model pow} + \text{LoRA pow}}$, validating that careful learning rate scaling is as critical as the choice of the PEFT method itself.

\paragraph{LoRA Rank.} Regarding the rank dimension, we challenge the notion that minimal ranks are sufficient for maximizing RL performance. While prior works suggest that even Rank=1 adapters can complete RLVR tasks effectively~\cite{mukherjee2025reinforcement}, our ablation across ranks 1, 8, 16, 32 reveals that relatively high ranks \textit{e.g.,} 16 and 32 yield superior results. Specifically, setting $r=1$ consistently underperformed higher-rank configurations. Given that the parameter overhead of LoRA is still negligible compared to the base model size, we advocate for avoiding extreme rank reduction; maintaining a moderate rank ensures sufficient expressivity to capture complex reasoning adjustments without compromising computational efficiency.

\subsection{Scaling on Stronger Models}
\label{sec:scale}


To verify the generalizability of our findings, we scale our evaluation to the 7B parameter regime using \texttt{DeepSeek-R1-Distill-Qwen-7B}. As summarized in Table~\ref{tab:task_avg_comparison_final}, the relative performance hierarchy observed in smaller models remains largely consistent at this larger scale. Both \textbf{DoRA} and \textbf{LoRA+} achieve an overall average accuracy of $55.0\%$, outperforming the standard LoRA baseline ($54.8\%$).  This consistent superiority suggests that the advantages of magnitude-direction decoupling (in DoRA) and optimized learning rate ratios (in LoRA+) are not mere artifacts of smaller model scales, but are intrinsic to the RLVR optimization landscape. Notably, DoRA maintains its lead across several challenging benchmarks, such as AMC ($83.1\%$) and AIME25 ($38.7\%$). These results reinforce our conclusion that for large-scale reasoning models, employing architecturally enhanced or optimization-aware awdapters is more effective than relying on the standard LoRA formulation.

\begin{table}[t]
    \centering
    \setlength{\tabcolsep}{5pt}
    \small
    \begin{tabular}{lccccccccccccc}
    \toprule
    \textbf{Methods} & \textbf{Avg.} & \multicolumn{2}{c}{\textbf{AIME24@32}}& \multicolumn{2}{c}{\textbf{AIME25@32}}& \multicolumn{2}{c}{\textbf{AMC@32}}& \multicolumn{2}{c}{\textbf{HMMT@32}}& \multicolumn{2}{c}{\textbf{MATH500@4}}& \multicolumn{2}{c}{\textbf{Minerva@4}}\\
     & &  Avg.&Pass&  Avg.&Pass&  Avg.&Pass&  Avg.&Pass&  Avg.&Pass& Avg.&Pass\\
    \midrule
    \multicolumn{2}{l}{\textbf{\textit{7B Model}}}& & & & & & & & & & & & \\
    \rowcolor{gray!10} LoRA& 54.8 & 48.3 & 73.3 & 35.9 & 73.3 & 81.4 & 97.5 & 22.0 & 53.3 & 80.9 & 94.8 & 27.2 & 40.1 \\
    \rowcolor{orange!10} DoRA& 55.0 & 45.8 & 70.0 & 38.7 & 66.7 & 83.1 & 97.5 & 23.7 & 56.7 & 80.1 & 93.2 & 25.9 & 39.0 \\
    \rowcolor{orange!10} MiSS& 53.4 & 42.5 & 63.3 & 36.4 & 70.0 & 77.9 & 97.5 & 22.2 & 60.0 & 80.8 & 93.2 & 26.5 & 39.0 \\
    \rowcolor{yellow!10} LoRA+& 55.5& 46.3& 73.3& 39.0& 70.0& 82.7& 100.0& 22.0& 66.7& 81.8& 94.6& 27.5& 40.8\\
    \bottomrule
    \end{tabular}
    \vspace{0.1cm}
    \caption{Comparison of accuracy (Avg.) and pass scores (Pass). All values are reported in percentages (std dev removed for clarity).}
    \label{tab:task_avg_comparison_final}
    \vspace{-0.4cm}
\end{table}

\section{Future Work}

\paragraph{Advanced Infrastructure and Scaling.}
While our current utilization of TRL facilitates broad PEFT compatibility, its limitations in large-scale distributed training necessitate a migration to more high-performance frameworks like VeRL. We plan to leverage this infrastructure upgrade to expand the scope of our evaluation beyond the DeepSeek-R1-Distill family and current short-horizon training schedules. Future experiments will rigorously test \textit{R1-Zero-like} cold-start paradigms and diverse model architectures under prolonged training steps to verify the stability and asymptotic performance of PEFT-RL at scale.

\paragraph{Mechanistic Interpretability of Adapter Dynamics.}
A critical objective is to decipher the theoretical underpinnings of our empirical findings. While we observe that structural variants (\textit{e.g.,} DoRA) align effectively with RLVR while SVD-based initializations (\textit{e.g.,} PiSSA) suffer from collapse, the precise mathematical reasons remain to be fully elucidated. We intend to conduct deeper investigations into the spectral evolution and optimization landscapes of these adapters, moving beyond empirical observation to establish a grounded theory of why specific structural biases are essential for the sparse, off-principal optimization nature of reinforcement learning.

\paragraph{Broader Frontiers and Deployment Stability.}
Finally, we aim to verify the universality of our findings by extending efficient RLVR into multimodal environments, multi-turn interactions, and asynchronous RL settings. Concurrently, we will address practical deployment challenges that are often overlooked, such as the numerical stability of weight merging and potential inconsistencies between training and inference phases. Addressing these engineering hurdles is essential for transitioning PEFT-RL from academic benchmarking to robust real-world application.

\section{Conclusion}

In this work, we present the first systematic and large-scale evaluation of various Parameter-Efficient Fine-Tuning (PEFT) methodologies within the Reinforcement Learning with Verifiable Rewards (RLVR) paradigm. Our investigation across over 12 PEFT variants and multiple model scales yields several definitive insights that challenge the current reliance on standard LoRA. Specifically, we demonstrate that structural variants consistently outperform standard LoRA and can even surpass the performance of full-parameter fine-tuning. This highlights the importance of magnitude-direction decoupling in handling the complex policy shifts of RL. Conversely, we uncover the structural misalignment of SVD-informed initialization strategies like PiSSA and MiLoRA; we provide a mechanistic explanation showing how these methods' focus on principal components conflicts with RLVR’s intrinsic off-principal update dynamics, leading to training instability or collapse. Furthermore, our results identify an \textbf{expressivity floor}, where extreme parameter reduction methods (\textit{e.g.,} VeRA or Rank-1 adapters) create a structural bottleneck that severely limits reasoning plasticity.

In summary, this research provides a clear roadmap for navigating the PEFT-RL landscape. We advocate for the community to move beyond the default adoption of standard LoRA in favor of geometry-aware adapters like DoRA, which offer a superior balance of efficiency and reasoning capability.

\bibliography{example_paper}
\bibliographystyle{icml2026}


\end{document}